\def\year{2022}\relax
\documentclass[letterpaper]{article} 
\usepackage{aaai22}  
\usepackage[linesnumbered,ruled,vlined]{algorithm2e}
\usepackage{graphicx}
\usepackage{booktabs}
\usepackage{multirow}
\usepackage{float}  
\usepackage{subcaption}
\usepackage{caption}
\usepackage{times}  
\usepackage{helvet}  
\usepackage{courier}  
\usepackage[hyphens]{url}  
\usepackage{graphicx} 
\usepackage{natbib}  
\usepackage{caption} 
\usepackage{algpseudocode}
\usepackage{newfloat}
\usepackage{listings}
\usepackage{amsmath}
\usepackage{amsfonts}
\usepackage{amsthm}
\usepackage{hyperref} 
\usepackage{arydshln}
\newtheorem{definition}{Definition}

\usepackage{amssymb}
\DeclareCaptionStyle{ruled}{labelfont=normalfont,labelsep=colon,strut=off} 
\floatstyle{ruled}
\newfloat{listing}{tb}{lst}{}
\floatname{listing}{Listing}

\title{Local Search for Integer Quadratic Programming}
\author{
    Xiang He, Peng Lin, Shaowei Cai \thanks{Corresponding author}\\
}
\affiliations{
    Key Laboratory of System Software (Chinese Academy of Sciences) and State Key Laboratory of Computer Science, Institute of Software, Chinese Academy of Sciences, Beijing, China\\
    School of Computer Science and Technology, University of Chinese Academy of Sciences, Beijing, China\\

    \{hexiang, linpeng, caisw\}@ios.ac.cn
}

\usepackage{bibentry}

\begin{document}

\maketitle

\begin{abstract} 
Integer Quadratic Programming (IQP) is an important problem in operations research. Local search is a powerful method for solving hard problems, but the
research on local search algorithms for IQP solving is still on its early stage.
This paper develops an efficient local search solver for solving general IQP, called LS-IQCQP. 
We propose four new local search operators for IQP that can handle quadratic terms in the objective function, constraints or both.
Furthermore, a two-mode local search algorithm is introduced, utilizing newly designed scoring functions to enhance the search process.
Experiments are conducted on standard IQP benchmarks QPLIB and MINLPLIB, comparing LS-IQCQP
with several state-of-the-art IQP solvers.
Experimental results demonstrate that LS-IQCQP is competitive with the most powerful commercial solver Gurobi and outperforms other state-of-the-art solvers. 
Moreover, LS-IQCQP has established 6 new records for QPLIB and MINLPLIB open instances.

\end{abstract}

\section{Introduction}
Integer Quadratic Programming (IQP) problems are mathematical optimization problems where the objective function, constraints, or both include quadratic polynomial functions, with the variables required to take integer values. IQP is a generalization of Integer Linear Programming (ILP), extending ILP by incorporating quadratic terms into the objective function or constraints.

With its strong expressive power, IQP can accommodate the requirements of a variety of real-world scenarios, leading to an extensive range of practical applications. Many combinatorial optimization problems can be described using the IQP formulation, such as the quadratic assignment problem~\cite{8}, max-cut~\cite{89}, and maximum clique~\cite{22}. 
In the industrial domain, many optimization problems can also be modeled using IQP, including finance portfolio optimization~\cite{37}, quadratic network design~\cite{51}, telecommunications routing~\cite{52}, energy management~\cite{54}, automatic control~\cite{75} and so on.

Integer Quadratic Programming is NP-hard~\cite{npQP,npIQP}, making it challenging to solve.
There are two classes of methods of solving Integer Quadratic Programming: complete methods and incomplete methods. Complete methods aim to compute the exact optimal solution and prove its optimality, while incomplete methods focus on obtaining high-quality solutions within a reasonable time.
Nearly all of the complete solvers addressing IQP are based on the Branch-and-Bound algorithm~\cite{90}. This includes most state-of-the-art IQP solvers~\cite{mittelmann-plots}, such as the commercial solvers Gurobi~\cite{gurobi2022gurobi} and CPLEX~\cite{cplex}, as well as the academic solver SCIP~\cite{scip}.
Nonetheless, Branch-and-Bound algorithms require exponential time in the worst case, and unfortunately, their exponential behavior frequently appears in practice~\cite{furini2019qplib}.


Incomplete methods are usually implemented as local search algorithms, which play an important role in solving NP-hard combinatorial problems~\cite{largels,  conls3, largels2}.
These algorithms aim to find good solutions quickly and have shown significant effectiveness in solving pseudo-Boolean optimization~\cite{lei2021efficient} and integer linear programming~\cite{lin2023new}. However, research on local search algorithms for solving Integer Quadratic Programming is still on its early stages.

In practice, IQP instances vary widely due to their diverse properties. Many local search algorithms are devotedly tailored for specific IQP types with certain attributes.
For instance, MA~\cite{qubo3} and 
ACSIOM~\cite{qubo1} are developed for quadratic unconstrained binary problems.
CQP~\cite{66} and DQP~\cite{68} are designed for convex problems with linear constraints, while BQPD~\cite{49} and SQIC~\cite{63} are suitable for nonconvex problems with linear constraints.
There are several local search solvers for Nonlinear Programming (a problem class that includes IQP), such as CONOPT, IPOPT and SNOPT~\cite{conopt,ipopt,snopt}. Nevertheless, these solvers are dedicated to real variables and not to integer ones.
Knitro~\cite{knitro} periodically uses local search algorithms to assist its complete algorithm, but these local search algorithms only serve as an auxiliary method and must be dependent on complete algorithms to work.

This paper is dedicated to proposing an efficient local search solver for IQP with two main goals: 1. The solver can handle any type of IQP, whether the problem is convex or not, regardless of the presence of quadratic terms in the objective function, constraints or both. 2. The solver is standalone and can quickly find high-quality solutions for IQP without relying on any complete or other algorithms.
 
\subsection{Our Contributions}
In this work, we introduce LS-IQCQP, a novel local search solver designed for solving the IQP problem.

\begin{itemize}
\item We propose four new local search operators for general IQP, capable of handling both convex and non-convex problems, regardless of quadratic terms in the objective function, constraints or both. Additionally, a two-mode local search algorithm is introduced, utilizing newly designed scoring functions to enhance the search process.
\item Experiments are conducted on standard IQP benchmarks QPLIB and MINLPLIB, comparing LS-IQCQP with several state-of-the-art IQP solvers. Experimental results demonstrate the excellent performance of LS-IQCQP, showing it is competitive with the most powerful commercial solver, Gurobi, indicating a significant improvement in the field of local search solvers for IQP.
\item 
 LS-IQCQP has established \textbf{6 new records for QPLIB and MINLPLIB open instances}, further demonstrating its exceptional solving capabilities.
\end{itemize}

\section{Preliminary}\label{multimodel}

\subsection{Integer Quadratic Programming}

Integer Quadratic Program (IQP) is an optimization problem in which either the objective function, or some of the constraints, or both, are quadratic
functions. The problem has the general form as follows:
\begin{equation}
\label{eq1}
\begin{split}
    \text{Minimize} \quad & \frac{1}{2} \mathbf{x}^T \mathcal{Q}^0 \mathbf{x} + b^0 \mathbf{x} + q^0 \\
    \text{subject to} \quad &\frac{1}{2} \mathbf{x}^T \mathcal{Q}^i \mathbf{x} + b^i \mathbf{x} \leq d_i \quad i \in \mathcal{M}, \\
                            & l^j \leq \mathbf{x}_j \leq u^j \quad j \in \mathcal{N}, \\
                            & \mathbf{x} \in \mathbb{Z}^n \\
\end{split}
\end{equation}

\begin{itemize}
    \item $\mathcal{N} = \{1, \ldots, n\}$ is the set of indices of variables;
    \item $\mathcal{M} = \{1, \ldots, m\}$ is the set of indices of constraints ($\mathcal{Q}^i = 0$ when constraint is linear constraint);
    \item $\mathbf{x} = [x_j]_{j=1}^n$ is a finite vector of integer variables, the variables are restricted to only attain integer values.
    \item $\mathcal{Q}^i$ for $i \in \{0\} \cup \mathcal{M}$ are symmetric $n \times n$ real (Hessian) matrices;
    \item $b^i$, $d_i$ for $i \in \{0\} \cup \mathcal{M}$, and $q^0$ are, respectively, real(integer) $n$-vectors and  constants;
    \item $-\infty \leq l^j \leq u^j \leq \infty$ are the lower and upper bounds on each variable $x_j$ for $j \in \mathcal{N}$.
    \item Any problem with a maximization objective can be converted into a minimization problem by negating the objective function.
\end{itemize}
To better introduce the algorithm, additional notations will be supplemented below.
For any variable $x_j \in$ constraint $i$, the constraint $con_i$ can be expressed using  $x_j$ as follows: 
\begin{equation}\label{consdef}
 A \cdot {x_j}^2 + H(i,x_j) \cdot x_j + I(i,x_j) \le d_i
\end{equation}
where A is the constant coefficient of  ${x_j}^2$ in $con_i$, $H(i,x_j)$ is the coefficient polynomial of $x_j$ in $con_i$, and $I(i,x_j)$ is the polynomial that represents the terms in $con_i $ that do not contain $x_j$. 

We use $f$ to denote the objective function. 
The set of variables in the objective function is denoted as $V(f)$, and the set of variables in a constraint $con_i$ as $V(con_i)$.
For any variable  $x_j \in V(f)$, let the polynomial  $\Theta({x_j})$ represent the terms in the objective function that involve $x_j$:
\begin{equation}\label{objdef}
\Theta({x_j}) = W \cdot {x_j}^2 + K(x_j) \cdot x_j 
\end{equation}
where $W$ is the constant coefficient of  ${x_j}^2$ in objective function, $K(x_j)$ is the coefficient polynomial of $x_j$ in the objective function.

A complete assignment (assignment for short) \(\alpha\) for an IQP instance \(F\) is a mapping that assigns to each variable an integer, and \(\alpha(x_j)\) denotes the value of \(x_j\) under \(\alpha\).
The value of the objective function under \(\alpha\) is denoted as $obj(\alpha)$.
Also, We denote \( \alpha(H(i,x_j)) \), \( \alpha(I(i,x_j)) \), and \( \alpha(K(x_j)) \) 
$\alpha (\Theta(x_j))$
as the values of these polynomials under assignment \( \alpha \), respectively.
An assignment \(\alpha\) satisfies the constraint $con_i$ if $ con_i(\alpha) \le d_i$, otherwise the constraint is violated. 
An assignment \(\alpha\) is \textbf{feasible} if and only if it satisfies all constraints in \(F\).
In the remaining part of the paper,  $\alpha$ means the current assignment, unless otherwise stated.



\section{Operator for Feasibility}\label{newoperator}

An operator defines how to modify variables to generate new assignments. When an operator is instantiated with a variable, it produces an operation. A local search algorithm progressively takes operations to generate new assignments and tracks the best assignment obtained. 

In this section, we propose a new operator {\it \textbf{quadratic satisfying move}} for violated quadratic constraints.
It considers modifying the value of variables in violated constraints towards making them satisfied.


\begin{definition}
The  {\it quadratic satisfying move} operator, denoted as $move_{sat}(x_j,con_i, \alpha)$,  takes an assignment $\alpha$ and assigns an integer variable $x_j$ to the threshold value making constraint $con_i$ satisfied, where $con_i$ is a violated constraint containing $x_j$.
\end{definition}
Note that for any variable \( x_j \in con_i \), \( con_i \) is given by \( A \cdot {x_j}^2 + H(i,x_j) \cdot x_j + I(i,x_j) \le d_i \) as in Formula~(\ref{consdef}).
We consider modifying the value of \( x_j \) while keeping any other variables fixed as constants.
There are two possible forms for \(con_i\) with respect to \(x_j\): it is linear  if \(A = 0\); otherwise, it is quadratic.
We will discuss these two cases individually below.
\subsection{I. Linear Form}

For any variable $x_j \in con_i$ with a zero quadratic coefficient \(A = 0\), the constraint $con_i$ is given by: $H(i,x_j) \cdot x_j + I(i,x_j) \le d_i$.
Under an assignment $\alpha$, 
we denote $\nu=\frac{d_i - \alpha(I(i,x_j))}{\alpha(H(i,x_j))}$, $\alpha(x^{\text{new}}_j)$ is the value of $x_j$ after performing $move_{sat}(x_j,con_i, \alpha)$ operator.
A $move_{sat}(x_j,con_i, \alpha)$ operator for {\it $x_j$} in $con_i$ is:

\begin{equation}
\alpha(x^{\text{new}}_j) =
\begin{cases}
\lceil \nu \rceil, & \text{if } H(I, x_j) > 0 , \\
\lfloor \nu \rfloor, & \text{if } H(I, x_j) < 0, \\
\alpha(x_j), & \text{otherwise}.
\end{cases}
\end{equation}

\subsection{II. Quadratic Form} 
For any variable \(x_j \in con_i\) with a non-zero quadratic coefficient \(A \neq 0\), the constraint $con_i$ is given by: $A \cdot {x_j}^2 + H(i,x_j) \cdot x_j + I(i,x_j) \le d_i$.
Under an assignment $\alpha$, 
we denote $\Delta=\alpha(H(I, x_j)^2) - 4 \cdot A \cdot (\alpha(I,x_j) - d_i)$. Depending on the following conditions, \( move_{sat}(x_j, con_i, \alpha) \) varies:

\begin{itemize}
    \item If $\Delta = 0$, the equation $A \cdot {x_j}^2 + H(i,x_j) \cdot x_j + I(i,x_j) - d_i = 0$ has a single  root $x_0$ with respect to $x_j$. If $x_0$ is an integer, then there is a $move_{sat}(x_j, con_i, \alpha)$ such that $\alpha(x_j^{\text{new}}) = x_0$. 
    
    \item If $\Delta > 0$, the equation $A \cdot {x_j}^2 + H(i,x_j) \cdot x_j + I(i,x_j) - d_i = 0$ has two roots $x_1$ and $x_2$ with respect to $x_j$, where $x_1 < x_2$, there are two $move_{sat}(x_j, con_i, \alpha)$:
    \begin{itemize}
    \item If \( A > 0 \), then the two \( move_{sat}(x_j, con_i, \alpha) \) are \( \alpha(x_j^{\text{new}}) = \lceil x_1 \rceil \) and \( \alpha(x_j^{\text{new}}) = \lfloor x_2 \rfloor \).
    \item If \( A < 0 \), then the two \( move_{sat}(x_j, con_i, \alpha) \) are \( \alpha(x_j^{\text{new}}) = \lfloor x_1 \rfloor \) and \( \alpha(x_j^{\text{new}}) = \lceil x_2 \rceil \).
\end{itemize}
    
    \item Otherwise, there is no $move_{sat}(x_j, con_i, \alpha)$.
\end{itemize}

Note that if \( con_i \) is an inequality, the new integer value of \( x_j \) will satisfy \( con_i \). However, if \( con_i \) is an equality, we must check whether the new integer value of \( x_j \) satisfies \( con_i \). If it does not, the \( move_{sat}(x_j, con_i, \alpha) \) operation will be aborted.


\section{Operators for Optimization}
For IQP, satisfying constraints is essential to find feasible solutions, while optimizing the objective function is crucial for achieving a (sub)optimal solution.
In this section, we propose 3 new operators for optimizing the objective function. 

\subsection{Inequality Exploration Move Operator}
We first introduce the {\it \textbf{inequality exploration move}} operator. It considers modifying the value of $x_j \in V(f)$ and the new value of $x_j$ is calculated based on the satisfied constraint $con_i$, where $con_i$ is an inequality that involves the $x_j$.

\begin{definition}
The inequality exploration move operator, denoted as \( move_{exp}(x_j, con_i, \alpha) \), takes an assignment $\alpha$ and assigns an integer variable \( x_j \in V(f) \) to a value that both maintains the satisfied state of the inequality constraint \( con_i \) involving \( x_j \) and maximize the decrease in the value of the objective function.
\end{definition}
$con_i$ is given by: $A \cdot {x_j}^2 + H(i,x_j) \cdot x_j + I(i,x_j) \le d_i$ with respect 
$x_j$ by Formula (\ref{consdef}).  If \(A = 0\),  the equation
$A \cdot {x_j}^2 + H(i,x_j) \cdot x_j + I(i,x_j) - d_i = 0$ has a single root $x_0$. If $A \neq 0$ and $\Delta > 0$, the equation has two roots $x_1$ and $x_2$, where $x_1 < x_2$.
The feasible domain of $x_j$ in $con_i$ is denoted as $\mathcal{D}$
, $\mathcal{D}$ is :

\begin{equation}
 \mathcal{D} =
\begin{cases}
[x_0,+\infty] & \text{if } A = 0  , \alpha(H(I, x_j)) < 0, \\
[-\infty, x_0] &  \text{if } A = 0  , \alpha(H(I, x_j)) > 0, \\
[x_1, x_2] &  \text{if } A > 0  , \Delta \neq 0 \\
[-\infty, x_1] \cup [x_2, +\infty] &  \text{if } A < 0  , \Delta \neq 0,
\end{cases}
\end{equation}


Recall that the terms in the objective function that involve \( x_j \) can be viewed as a quadratic function of \( x_j \), given by \( \Theta(x_j) = W \cdot x_j^2 + K(x_j) \cdot x_j \) according to Formula~(\ref{objdef}). If \( W \neq 0 \), let the axis of symmetry of this quadratic function be \( \xi = \frac{K(x_j)}{-2W} \). The inequality exploration move operator assigns an integer value \(\alpha(x^{\text{new}}_j)\) to \( x_j \) such that \(\alpha(x^{\text{new}}_j) \in \mathcal{D}\) and minimizes \(\Theta(x_j)\).

We determine a candidate value \( x^{\text{min}} \) based on the specified conditions of \( \text{con}_i \) and \( \Theta(x_j) \), where \( x^{\text{min}} \) is the value in \( \mathcal{D} \) that minimizes \( \Theta(x) \). 
This candidate value \( x^{\text{min}} \) is then used to find the nearest feasible integer within the domain \( \mathcal{D} \) to apply the \( move_{exp}(x_j, \text{con}_i, \alpha) \) operator. 
Specifically, $x^{\text{min}}$ is:

\begin{itemize}
    \item If \( W = 0 \), \( A = 0 \), and \( \alpha (H(I,x_j)) \cdot \alpha (K(x_j)) < 0 \),  
    then 
    \[
    x_{\text{min}} = x_0.
    \]

    \item If \( W = 0 \), \( A \neq 0 \), and \( \pm \infty \notin \mathcal{D} \), 
    then 
    \[
    x_{\text{min}} = \underset{x \in \{x_1, x_2\}}{\arg\min} \Theta(x).
    \]

    \item If \( W < 0 \) and \( \pm \infty \notin \mathcal{D} \), or if \( W > 0 \),
    \begin{itemize}
        \item If \( \xi \in \mathcal{D} \), then 
        \[
        x_{\text{min}} = \underset{x \in \{x_1, x_2, \xi\}}{\arg\min} \Theta(x).
        \]
        \item If \( \xi \notin \mathcal{D} \), then 
        \[
        x_{\text{min}} = \underset{x \in \{x_1, x_2\}}{\arg\min} \Theta(x).
        \]
    \end{itemize}

    \item Otherwise, there is no \( x^{\text{candidate}} \) for \(\alpha(x_j)\) operator.
\end{itemize}

Finally,  we get $move_{exp}(x_j, con_i, \alpha)$ operator as follows:
if \( \lceil x^{\text{min}} \rceil \in \mathcal{D} \), then \(\alpha(x^{\text{new}}_j) = \lceil x^{\text{min}} \rceil\); otherwise, \(\alpha(x^{\text{new}}_j) = \lfloor x^{\text{min}} \rfloor\).

\subsection{Equality Incremental Move Operator}

The {\it \textbf{equality incremental move}} operator is designed for \( x_j \in V(f) \) and a satisfied constraint \( con_i \), where \( con_i \) is an equality involving \( x_j \). 
In contrast to inequalities, for a quadratic equality, the feasible region of \( x_j \) consists of either a single point or two distinct points. It is challenging to find a value of \( x_j \) that keeps the equality satisfied while decreasing the value of the objective function. Therefore, the equality incremental move operator considers modifying both \( x_j \) and an auxiliary variable \( x' \) to decrease the value of the objective function, where \( x' \) is adjusted to maintain the satisfied state of $con_i$.

\begin{definition}

The equality incremental move operator, represented as \( move_{inc}(x_j, x', con_i, \alpha) \), modifies  a variable $x_j\in V(f)\cap V(con_i)$ and another variable $x'\in V(con_i)$, where $con_i$ is a satisfied equality constraint. Firstly, it increases or decreases the value of $x_j$ by 1 so that the value of $\Theta(x_j)$ is decreased, and it updates the value of $x'$ to make $con_i$ remain satisfied. 

\end{definition}

We first determine the updated value of \( x_j \). The value of \( x_j \) is adjusted by a simple incremental move according to \( \Theta(x_j) \) to make it decrease: \(\alpha( x^{\text{inc}}_j) = \alpha(x_j) - 1 \) or \(\alpha( x^{\text{inc}}_j) = \alpha(x_j) + 1 \), such that \( \alpha(\Theta(x^{\text{inc}}_j)) < \alpha(\Theta(x_j)) \).
Next, we determine the updated value of \( x’ \). For variable $x' \in con_i$, its new value is computed based on the updated \( x_j \) to ensure that \( con_i \) remains satisfied after adjusting both variables. Let the new assignment after the change in \( x_j \) from \( \alpha \) as \( \alpha_{x_j} \).
For the constraint:
\[
con_i: A \cdot {x'}^2 + H(i, x') \cdot x' + I(i, x') = d_i
\]
with respect to \( x' \) as defined by Formula (\ref{consdef}). We then modify the value of \( x' \) while keeping any other variables (besides \( x_j \)) fixed as constants. 
The set of integer roots of the equation: $A \cdot {x'}^2 + H(i, x') \cdot x' + I(i, x') - d_i = 0$ with respect to $x'$ under the assignment \( \alpha_{x_j} \) is computed and denoted as \( \mathcal{R} \). 
Let the polynomial \( \Theta(x_j, x') \) be the terms in the objective function that involve both \( x_j \) and \( x' \). 
For any root \( r_i \in \mathcal{R} \), let \( \alpha' \) denote the new assignment after changing the values of \( x_j \)  to \( x_j^{\text{inc}} \) and \( x' \) to \( r_i \) from the original assignment \( \alpha \). If:
$$
\alpha'(\Theta(x_j, x')) < \alpha(\Theta(x_j, x')),
$$
then there is a \( move_{inc}(x_j, x', con_i, \alpha) \) such that:
$$
\alpha(x_j^{\text{new}}) = \alpha(x_j^{\text{inc}}), 
\alpha(x'^{\text{new}}) = r_i.
$$

\subsection{Free Move Operator}
Considering a special case where a variable \( x_j \in V(f) \) has no constraints involving it. 
We denote $x_j$ as {\it \textbf{free variables}}. 
The {\it \textbf{free move}} operator, denoted as \( move_{free}(x_j, \alpha) \), is designed for such variables.
Recall that, $\Theta(x_j) = W \cdot x_j^2 + K(x_j) \cdot x_j .$
If \( W \neq 0 \), the axis of symmetry of this quadratic function is 
 $\xi = \frac{K(x_j)}{-2W} .$
It adjusts its value to minimize the objective function. Specifically, \( move_{free}(x_j, \alpha) \) is:

\begin{equation}
\alpha(x^{\text{new}}_j) =
\begin{cases}
\alpha(x_j) - 1, & \text{if } W = 0, K(x_j) > 0 \\
\alpha(x_j) + 1, & \text{if } W = 0,  K(x_j) < 0\\
\ \xi, & \text{if } W > 0\\
\alpha(x_j), & \text{otherwise}.
\end{cases}
\end{equation}

\section{Weighting Scheme and Score Function} 
In this section, we introduce techniques that help the local search algorithm perform effective search.

\subsection{Weighting Scheme}
The weighting scheme guides the search in a promising direction by assigning an additional property called weight (an integer) to constraints and the objective function. These weights are dynamically adjusted during the search and are used to compute the score functions.
We denote \( w(con_i) \) as the weight of constraint \( con_i \) and \( w(obj) \) as the weight of the objective function. Both \( w(con_i) \) and \( w(obj) \) are initially set to 1 and are updated dynamically as follows:

\begin{itemize}
\item  for violated constraint $con_i$, $w(con_i) := w(con_i) + 1$;
\item if $obj(\alpha)>obj^*$ and  $w(obj) < \zeta$, then $w(obj) := w(obj) + 1$, where $obj^*$ is objective value under best found solution, $\zeta$ limits the maximum value that objective function weight can get.
\end{itemize}

\subsection{Scoring Function}
The scoring function is a crucial component of local search algorithms, used to select best operations. We design the scoring function based on the feasibility of the current solution. For infeasible solutions, the scoring function assesses whether there is a change in the state of the constraints and the objective function. For feasible solutions, we evaluate how significant these changes are.
The score consists of two components: constraint score and objective function score. Given the weights for constraints and the objective function, the scoring function for an operation \( op \) that changes an assignment \(\alpha\) to \(\alpha'\) is designed as follows:

The constraint score measures the change in the total penalty of constraints.
Specifically,
if $\alpha$ is infeasible, any violated constraint under $\alpha$ and $\alpha'$ incurs a penalty of  $w(con_i)$.
if $\alpha$ is feasible, the penalty of all constraints under $\alpha$ is $p(con_i, \alpha) =0$,
any violated constraint under $\alpha'$ incurs a penalty of 
$p(con_i, \alpha') = w(con_i) \cdot \left| d_i - \left( \frac{1}{2} \mathbf{x}^T \mathcal{Q}^i \mathbf{x} + b^i \mathbf{x} \right) \right|$.

\begin{definition} [Constraint Score]
The constraint score for an operation $op$ is the decrease of the total penalty caused by performing 
$op$
: 
$$
Constraint \ Score(op) = \sum_{i=1}^{m} \left(p(con_i, \alpha) - p(con_i, \alpha')\right)
$$

\end{definition}

The objective function score measures the change in its value.
Let \( \Delta_{obj} \) as change  \( (obj(\alpha) - obj(\alpha')) \). 
If \( \Delta_{obj} \neq 0 \), then sign(\( \Delta_{obj} \))  
$ = \frac{\Delta_{\text{obj}}}{|\Delta_{\text{obj}}|}$, otherwise, 
sign(\( \Delta_{obj} \)) $=0$.  

\begin{definition} [Objective Function Score]
The objective function score for an operation \text{op} is \( w(\text{obj})\) $\cdot$ sign(\( \Delta_{obj} \))  if \( \alpha \) is infeasible, and score is \( w(\text{obj}) \cdot \Delta_{\text{obj}} \) if \( \alpha \) is feasible.

\end{definition}
\begin{definition} [Total Score]
The score of an operation is the sum of the constraint score and the objective function score.
\end{definition}

An operation is \textbf{decreasing} if its score is positive, indicating that it moves the search in a promising direction. 

\section{Local Search Algorithm} 
Our algorithm primarily consists of two aims: Satisfying all violated constraints and optimizing the objective function.
To achieve this, we design an algorithm with a two-mode structure: the {\it Satisfying} mode and the {\it Optimization} mode, each associated with its specific operators.
In the {\it Satisfying} mode, the algorithm focuses on making violated constraints satisfied, while in the {\it Optimization} mode, the algorithm focuses on minimizing the objective function.

\begin{algorithm}[!t]		
\caption{Satisfying Mode} 
\label{Infeasible Mode}
    \While{$\exists$ violated constraints}{
	\If{$ {\exists}$ decreasing quadratic satisfying move operation in violated constraints }{
		$op =$  a decreasing quadratic satisfying move operation with the greatest score via BMS;

	}
        \Else{
            update weights by Weighting Scheme;
            
            $c=$ a random violated constraint;
            
            $op =$ a quadratic satisfying move operation with the greatest score in $c$ via BMS;
       }
       perform op to modify $\alpha$;
   }
\end{algorithm}

\begin{algorithm}[!t]
\caption{Optimization Mode} 
\label{Optimization Mode}
    \While{$\forall$ constraints is satisfied }{
        $CandOp =  \{ \text{all $move_{exp}$, $move_{inc}$, $move_{free}$ operations} \} $;
        
	\If{$ {\exists}$ decreasing operaton in  $CandOp$}{
		$op= $ a decreasing operaton with the greatest score picked via BMS; \;
	}
        \Else{
            update weights by Weighting Scheme;
            
            $v =$ a random variable $ \in V(f)$;
            
            $op= $ an operaton for $v$ with the greatest score in $CandOp$  via BMS;
       }
       perform op to modify $\alpha$;
   }

\end{algorithm}
The routine of our algorithm is described in Algorithm \ref{guaranteed}.
First, the best assignment of best solution $\alpha^*$ and the current assignment $\alpha$ are initialized (Line 1), with all variables set to 0. 
Then, the algorithm iteratively modifies $\alpha$ by performing operations on integer variables. The algorithm dynamically switches between two modes based on whether there are any violated constraints (Line 2--8).

If there exist violated constraints, then our algorithm enters Satisfying mode.
As described in Algorithm ~\ref{Infeasible Mode}, 
the algorithm first tries to find a decreasing quadratic satisfying move operation with the greatest score via BMS heuristic~\cite{cai2015balance} (Line 2--3).
Specifically, the BMS heuristic samples $t$  operations ($t$ is a parameter), and selects the decreasing one with the greatest score.
If no decreasing operation is found, it indicates that the algorithm may reach a local optimum.
The algorithm further tries to escape from the local optimum by updating the weights of the constraints and the objective function based on the weighting scheme. Then it randomly selects one of the violated constraints and chooses a quadratic satisfying move operation with the greatest score via BMS (Line 4--7).

Otherwise, our algorithm enters the Optimization mode. In this mode, the algorithm focuses on minimizing the objective function. The Optimization mode is described in Algorithm \ref{Optimization Mode}. In each iteration, all inequality exploration move operations for inequality constraints, equality incremental move operations for equality constraints, and free move operations for free variables are added to a candidate set $CandOp$ (Line 2). The algorithm first tries to find a decreasing operation in $CandOp$ with the greatest score via the BMS heuristic (Line 3--4). If it fails to find any decreasing operation, the weighting scheme is activated. Then it selects a random variable $v \in V(f)$ and chooses an operation for $v$ in $CandOp$ with the greatest score (Line 5--8).

\begin{algorithm}[!t]		
\KwIn{IQP instance $F$, cutoff time {\it cutoff}}
\KwOut{Best found solution $obj(\alpha^*)$ or NA}
\caption{Local Search Algorithm} 
\label{guaranteed}
    $\alpha^* := \emptyset$,
    $\alpha :=$ an initial complete assignment\; 
    
    \While{time not exceeds $ cuttoff $}{
	\If{$\alpha$ is feasible and $obj(\alpha) < obj^*$}{
		$\alpha^* := \alpha,\; obj^* := obj(\alpha)$\;
	}
	\If{$ {\exists}$ violated constraints }{
		Enter {\it Satisfying Mode} and perform corresponding operation;
	}
 	\Else{
		Enter {\it Optimization Mode} and perform corresponding operations;
	}
   }
    \lIf {$\alpha^*$ is feasible}{\Return ($obj^*,\ \alpha^*$)}
    \lElse{\Return NA}
\end{algorithm}

\section{Experiments}


\label{experiments}
In this section, we evaluate the performance of our solver, LS-IQCQP, against state-of-the-art IQP solvers on standard IQP benchmarks. 
Ablation experiments are conducted to assess the effectiveness of the proposed strategies. 
We also conduct experiments to verify the stability of our solver.
Notably, LS-IQCQP has set new records for 6 open instances.

\subsection{Experiment Implementation and Setup:}
LS-IQCQP is programmed in C++, compiled by g++ with '-O3' option.
All experiments are carried out on a server with AMD EPYC 9654 CPU and 2048G RAM under the system Ubuntu 20.04.4.
There are 2 parameters in the solver:
$t$ the number of samples for the BMS heuristic,
$\zeta$ for the Upper Limit of objective function weight.
The parameters are tuned according to our preliminary experiments and suggestions from the literature, and are set as follows: $t=100$. $\zeta=100$.

\subsection{Competitors}
We compare LS-IQCQP with 4 state-of-the-art IQP solvers. The binaries for each competitor are downloaded from their respective websites and are run with default settings.
  
\begin{itemize}
\item \textbf{Gurobi}~\cite{gurobi2022gurobi}: The most powerful commercial IQP solvers. 
We use both its exact and heuristic versions, denoted by \textbf{Gurobi\_exact} and \textbf{Gurobi\_heur}, respectively (version 10.0.0).

\item \textbf{SCIP}~\cite{scip}: One of the fastest academic solvers for IQP (version 8.1.0).
\item \textbf{Cplex}~\cite{cplex}: A famous commercial IQP solver to solve IQP models (version 22.1.0).

\item \textbf{Knitro}~\cite{knitro}: A powerful mixed-integer nonlinear programming solver renowned for its advanced metaheuristics and local search algorithms (version 14.1).

\end{itemize}
 These solvers showcase different characteristics for IQP solving:  \textbf{Gurobi} is the best-performing IQP solver and \textbf{ranks first} in IQP competitions, as noted in~\cite{mittelmann-plots}. \textbf{SCIP} is the best open-source academic solver for IQP problems, also highlighted in~\cite{mittelmann-plots}. \textbf{Cplex}, although not participating in the IQP competition, is widely used and known for its strong performance in complex IQP models. Lastly, \textbf{Knitro} is the only solver we found that is both free for academic use and employs local search algorithms, while being capable of solving any type of IQP.


\subsection{Benchmarks}
Experiments are carried out with 2 benchmarks of the mainstream dataset for IQP. 
\begin{itemize}
\item \textbf{QPLIB}~\cite{furini2019qplib}: a  standard library of quadratic programming instances.
We select 137 instances where all variables are integers.
\item \textbf{MINLPLIB}~\cite{bussieck2003minlplib}: A library of mixed-integer and continuous nonlinear programming instances. 
We include 84 quadratic programming instances with integer variables, excluding those already present in QPLIB.
\end{itemize}
For each benchmark, instances are further categorized into 4 types:
1.\textbf{QUBO:} No constraints, quadratic objectives.
2.\textbf{LCQP:} Linear constraints only, quadratic objectives.
3.\textbf{QCLP:} Quadratic constraints, linear objectives.
4.\textbf{QCQP:} Quadratic constraints and objectives.
\subsection{Evaluation Metrics}
There are 3 Metrics for comparison in experiment.
1.\textbf{Winning instances:}
denoted as \#win, the number of instances where a solver finds the best solution among all solutions output by tested solvers.
2.\textbf{Feasible instances:}
denoted as \#feas, the number of instances where a solver can find a feasible solution within this time limit.
3.\textbf{Solving time:}
the runtime comparison between LS-IQCQP and competitor solvers if both solvers can find a solution with the same value of the objective function.
For each instance, each solver is run on a single thread with time limits of 10, 60, and 300 seconds, as referenced in~\cite{lin2023new}. In each table, the best performance for each metric under each time limit is highlighted in bold. Additionally, the number of instances in each benchmark is denoted as \#inst.

\subsection{Comparison with State-of-the-Art IQP Solvers}
 
{\bf The ability to find feasible(\#feas) solutions:}
As shown in Table 1.
LS-IQCQP performs best on both QPLIB and MINLPLIB benchmarks. It consistently finds feasible solutions for all instances in these benchmarks within 10 seconds or more. This result confirms the capability of LS-IQCQP to obtain feasible solutions within reasonable time limits.

{\bf The ability to find high-quality(\#win) solutions:}
As shown in Table 1. LS-IQCQP consistently leads other solvers in total \#win across all time limits, demonstrating its competitive performance.
In the QPLIB dataset, LS-IQCQP stands out particularly in the QUBO, QCLP and QCQP categories, showcasing its advantage in handling specific types of problems. 
In the MINLPLIB dataset, LS-IQCQP also performs exceptionally well in the QCLP and QCQP categories. LS-IQCQP's stable and robust performance across various problem types and time limits demonstrates its potential as an efficient solver.
It is worth mentioning that LS-IQCQP outperforms Knitro in all settings, indicating a significant improvement in the field of local search solver for IQP.

\begin{figure}[h]
\centering
\includegraphics[width=1.0\linewidth]{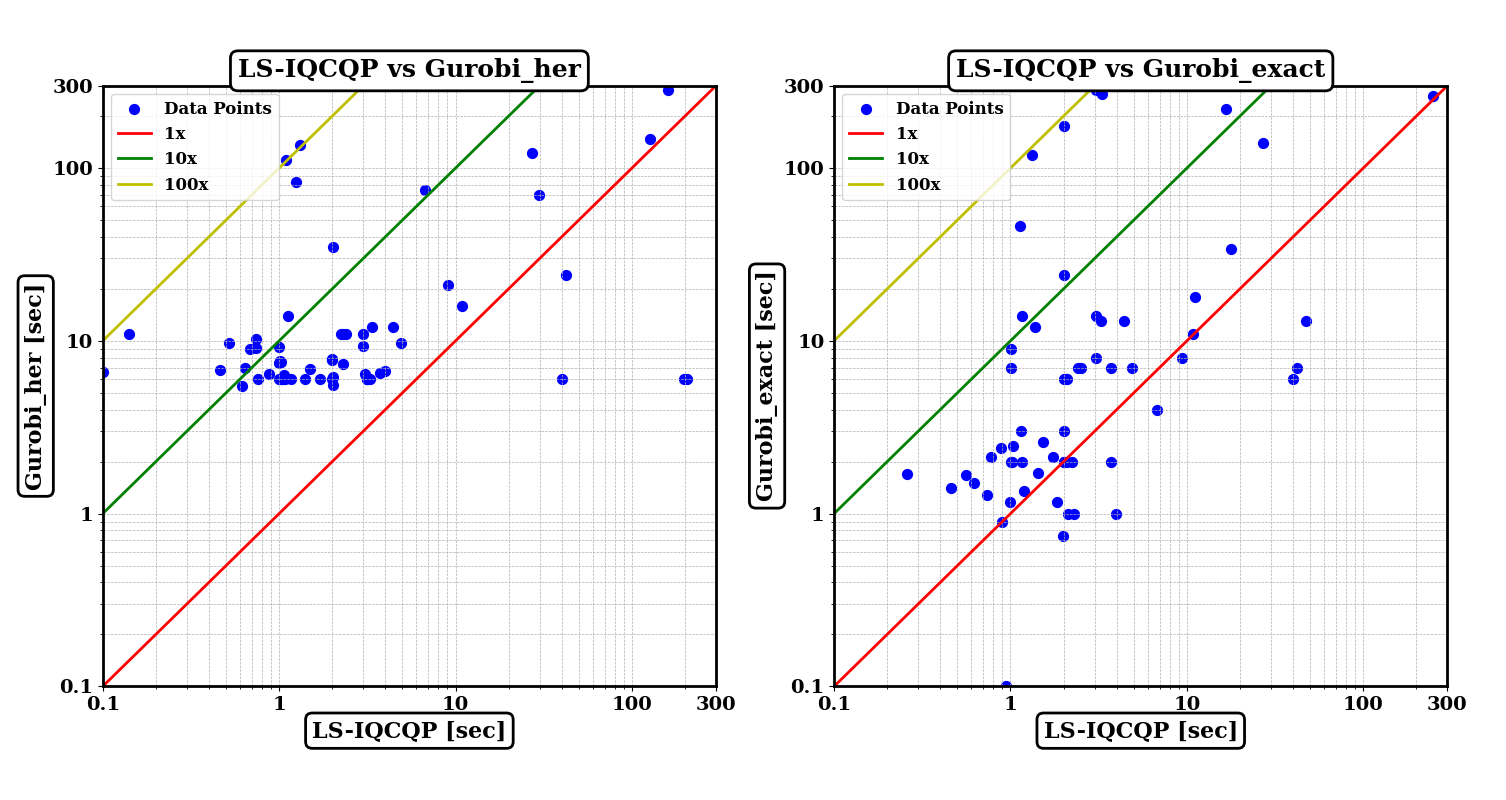}
\caption{
Running Time Comparision
}
\label{time}
\end{figure}

{\bf Solving time analysis:}
The solving time results are presented in Figure~\ref{time}. We compare LS-IQCQP with the top-performing solvers, Gurob\_exact and Gurobi\_heur, using a long 300-second time limit.
It can be observed that, for instances where the same solution is found, LS-IQCQP consistently obtains solutions in a shorter time. This demonstrates the rapid convergence speed of our solver.

\begin{table*}[t]
\centering
\label{win_table}
\renewcommand\arraystretch{0.6}
\setlength{\tabcolsep}{1.5mm}\scalebox{1.0}{
\begin{tabular}{c|c|c|cc|cc|cc|cc|cc|cc}
    \hline
    \multirow{2}*{Benchmark} & \multirow{2}*{Category} & \multirow{2}*{\textit{\#inst}} 
    & \multicolumn{2}{c|}{\textit{Gurobi\_exact}} 
    & \multicolumn{2}{c|}{\textit{Gurobi\_heur}} 
    & \multicolumn{2}{c|}{\textit{SCIP}} 
    & \multicolumn{2}{c|}{\textit{Cplex}} 
    & \multicolumn{2}{c|}{\textit{Knitro}} 
    & \multicolumn{2}{c}{\textit{LS-IQCQP}} \\
    \cline{4-15}
    & & & \textit{\#feas} & \textit{\#win} & \textit{\#feas} & \textit{\#win} & \textit{\#feas} & \textit{\#win} & \textit{\#feas} & \textit{\#win} & \textit{\#feas} & \textit{\#win} & \textit{\#feas} & \textit{\#win} \\
    \hline

\multicolumn{15}{c}{\textbf{Time Limit 10 Seconds}} \\
\hline
\multirow{4}{*}{QPLIB}
&QUBO &23 & \textbf{23} & 6 & \textbf{23} & 12 & \textbf{23} & 1 & \textbf{23} & 3 & \textbf{23} & 0 & \textbf{23} & \textbf{18} \\
&LCQP &99 & \textbf{99} & 56 & \textbf{99} & 73 & 87 & 23 & \textbf{99} & 30 & 80 & 10 & \textbf{99} & \textbf{74} \\
&QCLP &10 & \textbf{10} & 1 & \textbf{10} & 2 & \textbf{10} & 1 & 9 & 2 & \textbf{10} & 2 & \textbf{10} & \textbf{7} \\
&QCQP &5 & \textbf{5} & 0 & \textbf{5} & 1 & \textbf{5} & 0 & 4 & 0 & \textbf{5} & 0 & \textbf{5}  & \textbf{5} \\
\hdashline
\multirow{4}{*}{MINLPLIB}
&QUBO &19 & \textbf{19} & 6 & \textbf{19} & \textbf{13} & \textbf{19} & 2 & \textbf{19} & 2 & \textbf{19} & 0 & \textbf{19} & 9 \\
&LCQP &52 & \textbf{52} & 46 & \textbf{52} & \textbf{51} & 46 & 45 & \textbf{52} & 45 & 45 & 21 & \textbf{52} & \textbf{51} \\
&QCLP &2 & \textbf{2} & \textbf{2} & \textbf{2} & \textbf{2} & \textbf{2} & \textbf{2} & \textbf{2} & \textbf{2} & \textbf{2} & \textbf{2} &  \textbf{2} & \textbf{2} \\
&QCQP &11 & \textbf{11} & \textbf{11} & \textbf{11} & \textbf{11} & \textbf{11} & \textbf{11} & \textbf{11} & \textbf{11} & \textbf{11} & \textbf{11} & \textbf{11} & \textbf{11} \\ \hline
Total&  & 221 & \textbf{221} & 128 & \textbf{221} & 165 & 203 & 85 & 220 & 95 & 195 & 46 & \textbf{221} & \textbf{177} \\ \hline

\multicolumn{15}{c}{\textbf{Time Limit 60 Seconds}} \\
\hline
\multirow{4}{*}{QPLIB}
&QUBO &23 & \textbf{23} & 12 & \textbf{23} & 9 & \textbf{23} & 2 & \textbf{23} & 5 & \textbf{23} & 0 & \textbf{23} & \textbf{19} \\
&LCQP &99 & \textbf{99} & 64 & \textbf{99} & \textbf{74} & 98 & 26 & \textbf{99} & 37 & 87 & 12 & \textbf{99} & \textbf{74} \\
&QCLP &10 & \textbf{10} & 1 & \textbf{10} & \textbf{5} & \textbf{10} & 1 & \textbf{10} & 3 & \textbf{10} & 3 & \textbf{10} & 4 \\
&QCQP &5 & \textbf{5} & 1 & \textbf{5} & 1 & \textbf{5} & 0 & \textbf{5} & 0 & \textbf{5} & 0 & \textbf{5} & \textbf{5} \\
\hdashline
\multirow{4}{*}{MINLPLIB}
&QUBO &19 & \textbf{19} & 10 & \textbf{19} & \textbf{14} & \textbf{19} & 2 & \textbf{19} & 3 & \textbf{19} & 0 & \textbf{19} & 10 \\
&LCQP &52 & \textbf{52} & 46 & \textbf{52} & \textbf{52} & \textbf{52} & 45 & \textbf{52} & 45 & \textbf{45} & 21 & \textbf{52} & 51 \\
&QCLP &2 & \textbf{2} & \textbf{2} & \textbf{2} & \textbf{2} & \textbf{2} & \textbf{2} & \textbf{2} & \textbf{2} & \textbf{2} & \textbf{2} &  \textbf{2} & \textbf{2} \\
&QCQP &11 & \textbf{11} & \textbf{11} & \textbf{11} & \textbf{11} & \textbf{11} & \textbf{11} & \textbf{11} & \textbf{11} & \textbf{11} & \textbf{11} & \textbf{11} & \textbf{11} \\ \hline
Total&  & 221 & \textbf{221} & 147 & \textbf{221} & 168 & 220 & 89 & \textbf{221} & 106 & 202 & 49 & \textbf{221} & \textbf{176} \\ \hline

\multicolumn{15}{c}{\textbf{Time Limit 300 Seconds}} \\
\hline
\multirow{4}{*}{QPLIB}
&QUBO &23 & \textbf{23} & 13 & \textbf{23} & 9 & \textbf{23} & 5 & \textbf{23} & 9 & \textbf{23} & 0 & \textbf{23} & \textbf{20} \\
&LCQP &99 & \textbf{99} & 73 & \textbf{99} & \textbf{78} & \textbf{99} & 30 & \textbf{99} & 42 & 88 & 15 & \textbf{99} & 74 \\
&QCLP &10 & \textbf{10} & 2 &\textbf{10} & \textbf{6} & \textbf{10} & 1 & \textbf{10} & 3 & \textbf{10} & 5 & \textbf{10} & \textbf{6} \\
&QCQP &5 & \textbf{5} & 1 & \textbf{5} & 3 & \textbf{5} & 0 & \textbf{5} & 0 & \textbf{5} & 0 & \textbf{5} & \textbf{5} \\
\hdashline
\multirow{4}{*}{MINLPLIB}
&QUBO &19 & \textbf{19} & 11 & \textbf{19} & \textbf{15} & \textbf{19} & 5 & \textbf{19} & 8 & \textbf{19} & 0 & \textbf{19} & 9 \\
&LCQP &52 & \textbf{52} & 46 & \textbf{52} & \textbf{52} & \textbf{52} & 45 & \textbf{52} & 45 & 46 & 21 & \textbf{52} & 51 \\
&QCLP &2 & \textbf{2} & \textbf{2} & \textbf{2} & \textbf{2} & \textbf{2} & \textbf{2} & \textbf{2} & \textbf{2} & \textbf{2} & \textbf{2} &  \textbf{2} & \textbf{2} \\
&QCQP &11 & \textbf{11} & \textbf{11} & \textbf{11} & \textbf{11} & \textbf{11} & \textbf{11} & \textbf{11} & \textbf{11} & \textbf{11} & \textbf{11} & \textbf{11} & \textbf{11} \\ \hline
Total&  & 221 & \textbf{221} & 159 & \textbf{221} & 176 & \textbf{221} & 99 & \textbf{221} & 120 & 203 & 54 & \textbf{221} & \textbf{178} \\ \hline

\end{tabular}
}
\caption{Comparison of the number of \#win and \#feas instances with state-of-the-art IQP solvers}
\end{table*}

\vspace{-2ex}
\subsection{New Records to Open Instances}
In the QPLIB and MINLPLIB datasets, there are instances that remain open, indicating that the optimal solution has not yet been found. The current best-known solutions for each open instance are available on the respective QPLIB and MINLPLIB websites. These open instances represent some of the most challenging IQP problems to solve. \textbf{Remarkably, LS-IQCQP has established new best-known solutions for 6 open instances.} As shown in Table 2, these instances also include different types of IQP from QPLIB and MINLPLIB, demonstrating the powerful solving ability.

\begin{table*}[h]
\centering
\renewcommand\arraystretch{0.8}
\label{break}

\setlength{\tabcolsep}{1.8mm}\scalebox{1.0}{
\begin{tabular}{|c|c|c|c|c|c|c|c|}
\hline
 Instance Name & Benchmark& Category &\#var & \#cons & Objective  & Previous best & \textit{LS-IQCQP} \\  \hline
 QPLIB\_2036&QPLIB& QCQP & 324 &324 &Minimize  & -30590 & \textbf{-30660} \\
 QPLIB\_2096&QPLIB& QCQP & 300 &6925 &Minimize  & 7068000 & \textbf{7064664} \\
 chimera\_mgw-c16-2031-01&MINLPLIB& QUBO  & 2032 & 0 &Maximize& 1993 & \textbf{2006} \\
 chimera\_mgw-c16-2031-02&MINLPLIB& QUBO & 2032 & 0 &Maximize& 1996 & \textbf{2004} \\
 chimera\_selby-c16-01&MINLPLIB& QUBO & 2031 & 0 &Maximize& 739.4 & \textbf{741.1} \\
 chimera\_selby-c16-02&MINLPLIB& QUBO & 2031 & 0 &Maximize& 733.9 & \textbf{738.1} \\
 \hline
\end{tabular}
}
\caption{LS-IQCQP sets new records for 6 open instances. \#var and \#cons denote the number of variables and constraints, respectively. Minimize refers to a minimization problem, while Maximize refers to a maximization problem.}
\end{table*}

\subsection{Effectiveness of Proposed Strategies}

To evaluate the effectiveness of our proposed strategies, we develop three modified versions of LS-IQCQP. Experiments are conducted with a time limit of 300 seconds. The modifications and corresponding versions are as follows:

\begin{itemize}
\item Remove \textit{Inequality exploration move}: version $v_{no\_exp}$
\item Remove \textit{Equality incremental move}: version $v_{no\_inc}$
\item Remove \textit{Free move}: version $v_{no\_free}$
\end{itemize}

\begin{table}[h]
    \centering
    \label{com_table}
    \renewcommand\arraystretch{0.5}
    \setlength{\tabcolsep}{1.0pt}\scalebox{1.0}{
        \begin{tabular}{@{}lcccccc@{}}
            \toprule
            \multirow{2}{*}{Benchmark}  & \multicolumn{2}{c}{$v_{no\_exp}$} & \multicolumn{2}{c}{$v_{no\_inc}$} 
             & \multicolumn{2}{c}{$v_{no\_free}$} 
            \\ \cmidrule(lr){2-3} \cmidrule(lr){4-5} \cmidrule(lr){6-7}
            &  $\#better$ & $\#loss$ & $\#better$ & $\#loss$  & $\#better$ & $\#loss$\\ \midrule
            QPLIB & \textbf{65} & 2 & \textbf{40} & 5 & \textbf{23} & 4 \\ 
            MINLPLIB & \textbf{45} & 5 & \textbf{9} & 1 & \textbf{19} & 2 \\ 
            Total & \textbf{110} & 7 & \textbf{49} & 6 & \textbf{42} & 6 \\
            \bottomrule
        \end{tabular}
    }
     \caption{Comparing LS-IQCQP with modified versions. The number of instances where LS-IQCQP finds better(\#better) solutions and worse(\#loss) solutions are presented.}
\end{table}


The results of ablation experiment are presented in Table 3 and confirm the effectiveness of the proposed strategies.
\subsection{Stability Experiments}
{\bf Repetitive running with randomness}
To examine the stability of LS-IQCQP, we execute the algorithm 10 times using seeds from 1 to 10 for each instance. 
Experimental results show that over 80\% of instances have a coefficient of variation less than 0.1, demonstrating that LS-IQCQP exhibits stable performance.



 {\bf Sensitivity analysis on the parameters}
To examine the sensitivity of parameters in LS-IQCQP, we test the algorithm with various parameter settings. Results are in Appendix C and show that these variations do not significantly affect the algorithm's overall performance.





\section{Conclusion and Future Work}
In this paper, we focus on local search algorithms for IQP. We propose 4 novel operators and a two-mode algorithm with new scoring functions to enhance the search process.  Experiments show that the performance of our solver is outstanding, setting 6 new records for open instances.
Future work aims to combine our solver with other solvers to  prune the search space and speed up the search process.
\label{conclusion}

\bibliography{lipics-v2019-sample-article}


\end{document}